\documentclass[10pt,twocolumn,letterpaper]{article}

\usepackage{iccv}
\usepackage{times}
\usepackage{epsfig}
\usepackage{graphicx}
\usepackage{amsmath}
\usepackage{amssymb}
\usepackage{multirow}
\usepackage{tabularx}
\usepackage[numbers]{natbib}
\usepackage{booktabs}
\usepackage{subcaption}
\usepackage[accsupp]{axessibility}  

\usepackage[pagebackref=true,breaklinks=true,letterpaper=true,colorlinks,bookmarks=false,linkcolor=black]{hyperref}

\usepackage{algorithm}
\usepackage{algpseudocode}

\iccvfinalcopy 

\def\iccvPaperID{11381} 

\ificcvfinal\fi

\begin{document}

\title{Robust Evaluation of Diffusion-Based Adversarial Purification}

\author{Minjong Lee\\
CSED POSTECH\\
{\tt\small minjong.lee@postech.ac.kr}
\and
Dongwoo Kim\\
CSED, GSAI POSTECH\\
{\tt\small dongwoo.kim@postech.ac.kr}
}

\maketitle
\ificcvfinal\thispagestyle{empty}\fi

\begin{abstract}
We question the current evaluation practice on diffusion-based purification methods. Diffusion-based purification methods aim to remove adversarial effects from an input data point at test time. The approach gains increasing attention as an alternative to adversarial training due to the disentangling between training and testing.
Well-known white-box attacks are often employed to measure the robustness of the purification. However, it is unknown whether these attacks are the most effective for the diffusion-based purification since the attacks are often tailored for adversarial training.
We analyze the current practices and provide a new guideline for measuring the robustness of purification methods against adversarial attacks. Based on our analysis, we further propose a new purification strategy improving robustness compared to the current diffusion-based purification methods.
\end{abstract}

\section{Introduction}
\label{Introduction}

Adversarial attacks~\citep{Madry2017TowardsDL, Croce2020ReliableEO} can cause deep neural networks (DNNs) to produce incorrect outputs by adding imperceptible perturbations to inputs. While various adversarial defenses have been proposed, adversarial training \cite{zhang2019theoretically, Gowal2021ImprovingRU} has shown promising results in building robust DNNs. Since adversarial training feeds the model both normal and adversarial examples during training time, one needs to pre-determine which attack method is used to generate the adversarial examples. On the other hand, adaptive test-time defense~\citep{kang2021stable, ho2022disco} has recently gained increasing attention since it adaptively removes the adversarial effect at test time without adversarial training. \textit{Adversarial purification}~\citep{Samangouei2018DefenseGANPC, Nie2022DiffusionMF}, one of the adaptive test-time defenses, uses generative models to restore the clean examples from the adversarial examples.

Diffusion-based generative models~\citep{Ho2020DenoisingDP, Song2020ScoreBasedGM} has been suggested as a potential solution for adversarial purification~\citep{Nie2022DiffusionMF, Wang2022GuidedDM}. Diffusion models learn transformations from data distributions to well-known simple distributions such as the Gaussian and vice versa through forward and reverse processes, respectively. When applied to the purification, the forward process gradually adds noise to the input, and the reverse process gradually removes the noises to uncover the original image without imperceptible adversarial noise.
With a theoretical guarantee, the recent success of DiffPure~\citep{Nie2022DiffusionMF} against many adversarial training methods shows the potential of using diffusion processes for improving the robustness against adversarial attacks.

Evaluating the robustness of adaptive test-time defenses is, however, known to be difficult due to their complex defense algorithms and properties. 
\citet{croce2022evaluating} shows that finding worse-case perturbation is important to measure the robustness of the defenses.
Randomness and iterative calls of adaptive test-time defenses, however, make their gradients obfuscated. Therefore, the gradient-based attack methods~\citep{Madry2017TowardsDL, Croce2020ReliableEO} might be inappropriate for measuring the robustness under such obfuscation. 
New algorithms, such as Backward Pass Differentiable Approximation (BPDA)~\citep{obfuscated-gradients}, and additional recommendations~\citep{croce2022evaluating} have been proposed to evaluate their robustness accurately. However, it is unclear whether these algorithms and recommendations can still be used to evaluate diffusion-based purification.

In the first part of this work, we analyze the existing evaluation methods for diffusion-based purification.
We find that the adjoint method, often used to compute a full gradient of the iterative process, relies on the performance of an underlying numerical solver. Tailored to the diffusion models, we propose a surrogate process, an alternative method to approximate the gradient from the iterative procedure and show the strong robustness in recent work can be weaker than claimed with the surrogate process. We then compare two gradient-based attack methods, AutoAttack~\citep{Croce2020ReliableEO} and PGD~\citep{Madry2017TowardsDL}, with the surrogate process and find that PGD is more effective for the diffusion-based purification. To this end, we propose a practical recommendation to evaluate the robustness of diffusion-based purification.

In the second part of this work, we analyze the importance of the hyperparameter for successive defenses with purification. 
The diffusion models are trained without adversarial examples. Thus, proper validation of hyperparameters is impossible in general.
Instead, we empirically analyze the influence of different hyperparameter selections from the attacker's and defender's perspectives. 
Based on our analysis, we propose a gradual noise-scheduling for multi-step purification. 
We show that our defense strategy highly improves robustness compared to the current diffusion-based purification methods under our proposed evaluation scheme.

We summarize our contributions as follows:
\begin{itemize}
    \item We analyze the current evaluation of diffusion-based purification and provide a recommendation for robust evaluation.
    \item We investigate the influence of hyperparameters on the robustness of diffusion-based purification.
    \item We propose a gradual noise-scheduling strategy for diffusion-based purification, improving the robustness of diffusion-based purification.
\end{itemize}

\section{Preliminary}
\label{Preliminary}
We provide the background on the adversarial attack, diffusion models, and adversarial purification in this section.

\subsection{Adversarial Attacks}
Adversarial attacks aim to manipulate or trick machine learning models by adding imperceptible perturbations to input data that can cause the model to misclassify or produce incorrect outputs. The adversarial attacks can be categorized into black-box, grey-box, and white-box attacks. The black-box attack assumes that the attacker knows nothing about the internal structure of the classifier and defender. The white-box attack assumes that the attacker can obtain any information about the defender and the target classifier, including the architecture and parameter weights. The grey-box lies between the white- and black-box attacks, where the attacker partially knows the target model. In this work, we only focus on the performance of purification in the white-box attack since the white-box attack is the most difficult to defend from the defender's perspective.

The Projected Gradient Descent (PGD)~\citep{Madry2017TowardsDL} method is a common white-box attack. PGD is a gradient-based attack that iteratively updates an adversarial example using the following rule
\begin{equation}
\mathbf{x}_{i + 1} = \Pi_{\mathcal{X}} \left(\mathbf{x}_i + \alpha_i \text{sign} \nabla_\mathbf{x} \mathcal{L}(f_\phi(\mathbf{x}), y) |_{\mathbf{x}=\mathbf{x}_i}\right),
\end{equation}
where $f_\phi$ represents a classifier, and $\Pi_\mathcal{X}$ indicates a projection operation onto $\mathcal{X}$. PGD can only be applied for the differentiable defense methods. For non-differentiable defense methods, the Backward Pass Differentiable Approximation (BPDA)~\cite{obfuscated-gradients} is widely used, which computes the gradient of the non-differentiable function by using a differentiable approximation. Expectation over Transformation (EOT)~\citep{Athalye2017SynthesizingRA} can be additionally employed for randomized defenses, which optimizes the expectation of the randomness. AutoAttack~\citep{Croce2020ReliableEO} is an ensemble of four different types of attacks. In this work, we measure the robustness of purification methods against these attack methods. 

\subsection{Diffusion Models}
Recently, diffusion-based models~\citep{Ho2020DenoisingDP, Song2020ScoreBasedGM} have gained increasing attention in generative models. Unlike the VAEs and GANs, the diffusion-based models produce samples by gradually removing noise from random noise. The training of diffusion-based models consists of two processes, the forward process, and the reverse denoising process.
The forward process adds Gaussian noise over $T$ steps to the observed input $\mathbf{x}_0$ with a predefined variance scheduler $\beta_t$, whose joint distribution is defined as 
\begin{equation}
    q(\mathbf{x}_{1:T} \vert \mathbf{x}_0) = \prod^T_{t=1} q(\mathbf{x}_t \vert \mathbf{x}_{t-1}),
\end{equation}
where $q(\mathbf{x}_t \vert \mathbf{x}_{t-1})$ is a Gaussian transition kernel from $\mathbf{x}_{t-1}$ to $\mathbf{x}_t$ 
\begin{equation}
    q(\mathbf{x}_{t} | \mathbf{x}_{t-1}) := \mathcal{N}(\mathbf{x}_{t}; \sqrt{1 - \beta_t} \mathbf{x}_{t-1}, \beta_t\textbf{I}).
\end{equation}
The reverse process denoises the random noise $\mathbf{x}_{T}$ over $T$ times, whose joint distribution is defined as
\begin{equation}
    p_\theta(\mathbf{x}_{0:T}) = p(\mathbf{x}_T) \prod^T_{t=1} p_\theta(\mathbf{x}_{t-1} | \mathbf{x}_t).
\end{equation}
The transition distribution from $\mathbf{x}_t$ to $\mathbf{x}_{t-1}$ is often modeled by Gaussian distribution
\begin{equation}
    p_\theta(\mathbf{x}_{t-1} | \mathbf{x}_t) = \mathcal{N}(\mathbf{x}_{t-1}; \boldsymbol{\mu}_\theta(\mathbf{x}_t, t), \sigma^2_t \textbf{I}),
\end{equation}
where $\sigma_t$ is a variance, and $\boldsymbol{\mu}_\theta$ is a predicted mean of $\mathbf{x}_{t-1}$ derived from a learnable denoising model $\boldsymbol{\epsilon}_\theta$. The denoising model is often trained by predicting a random noise at each time step via following objective
\begin{equation}
L(\theta) = \mathbb{E}_{t, \mathbf{x}_0, \boldsymbol{\epsilon}} \Big[\|\boldsymbol{\epsilon} - \boldsymbol{\epsilon}_\theta(\mathbf{x}_t, t)\|^2 \Big], \\
\end{equation}
where $\boldsymbol\epsilon$ is a Gaussian noise, i.e., $\boldsymbol\epsilon \sim \mathcal{N}(0, \textbf{I})$. 
The model $\boldsymbol{\epsilon}_\theta$ takes the noisy input $\mathbf{x}_t$ and the time step $t$ to predict the actual noise $\boldsymbol\epsilon$ at time $t$.
In the Denoising Diffusion Probabilistic Model (DDPM)~\citep{Ho2020DenoisingDP}, the reverse denoising process is performed over $T$ steps through random sampling, resulting in a slower generation of samples compared with GANs and VAEs.

Based on the fact that the multiple denoising steps can be performed at a single step via a non-Markovian process, \citet{Song2020DenoisingDI} proposes a new sampling strategy, which we call Denoising Diffusion Implicit Model (DDIM) sampler, to accelerate the reverse denoising process. In this work, we compare the performances of DDPM and DDIM samplers in the diffusion-based purification approach.

\subsection{Adversarial Purification}
Adversarial purification via generative models is a technique used to improve the robustness of machine learning models against adversarial attacks~\citep{Samangouei2018DefenseGANPC}. The idea behind this technique is to use a generative model to learn the underlying distribution of the clean data and use it to purify the adversarial examples.

Diffusion-based generative models can be used as a purification process if we assume that the imperceptible adversarial signals as noise~\citep{Nie2022DiffusionMF}.
To do so, the purification process adds noise to the adversarial example via the forward process with $t^*$ steps, and it removes noises via the denoising process. The choice of the number of forward steps $t^*$ is essential since too much noise can remove the semantic information of the original example, or too little noise cannot remove adversarial perturbation. 
In theory, as we add more noise to the adversarial example, the distributions over the noisy adversarial example and the true example become close to each other~\citep{Nie2022DiffusionMF}. Therefore, the denoised examples are likely to be similar.

\section{Evaluation for Diffusion-Based Purification}
\label{sec:evaluation}
In this section, we first review the current practices in evaluating diffusion-based purification methods. We then curate three research questions to address their potential limitations and provide our answers to these questions through empirical evaluations.

\subsection{Current Practices and Research Questions}
Evaluation of the diffusion-based purification against gradient-based white-box attacks is non-trivial due to many function calls on the denoising process. Multiple function calls in the denoising step often require an impractical amount of memory, making it unfeasible to compute the gradient of the full defense process. Because of this problem, most defenses~\citep{Yoon2021AdversarialPW, Wang2022GuidedDM, ho2022disco} consider BPDA the strongest adaptive white-box attack in the current practice since it does not rely on the gradients of defense methods. However, the vulnerability of diffusion-based purification on white-box attacks has yet to be fully identified. The importance of testing in adaptive white-box attacks of purification has been recognized only recently by the work of DiffPure~\citep{Nie2022DiffusionMF}.

DiffPure calculates the full gradients of their defense process using an adjoint method. The adjoint method is employed to avoid the extensive use of memory while obtaining the full gradient. DiffPure is evaluated on AutoAttack, a de facto evaluation method in adversarial training. Although their evaluation framework is more robust than the previous work, the design choices of their evaluation still raise questions since 1) the adjoint method relies on the performance of an underlying {numerical} solver~\citep{Zhuang2020AdaptiveCA}, and 2) there is no comprehensive comparison between different attacks using the full gradient.

Based on our observation, we carefully curate the following three research questions to address the robustness of the current evaluation framework in diffusion-based purification:
\begin{itemize}
    \item \textbf{RQ1.} Is the adjoint method the best way to generate adversarial examples with full gradients? Is there any alternative to the adjoint method?
    \item \textbf{RQ2.} Is AutoAttack still better than the other gradient-based attacks, such as PGD, when the alternative is available?
    \item \textbf{RQ3.} Is BPDA still more effective than the best combination of full-gradient attacks?
\end{itemize}
In the next section, we re-evaluate the existing purification methods to answer these questions.

\subsection{Experimental Results \& Analysis}
\label{sec:results_and_analysis}

We evaluate the performance of three diffusion-based purification methods: ADP\footnote{Although ADP uses a score-based model and Langevin dynamics, since the concept is similar to the diffusion model, we consider ADP diffusion-based purification.}~\citep{Yoon2021AdversarialPW}, DiffPure~\citep{Nie2022DiffusionMF}, GDMP~\citep{Wang2022GuidedDM}. We additionally evaluate two non-diffusion-based adaptive test-time defenses: SODEF~\citep{kang2021stable}, and DISCO~\citep{ho2022disco} to address whether our findings still hold for the non-diffusion-based purification methods.
We evaluate their robustness on CIFAR-10 against three gradient-based attacks, including PGD, BPDA, and AutoAttack, with a maximum attack strength of $\ell_\infty (\epsilon = 8/255)$. 
A comprehensive description of evaluation configurations is provided in \autoref{app:defense_and_evaluation}.

\paragraph{Surrogate process and its gradient.}
The adjoint method can compute the exact gradient in theory, but in practice, the adjoint relies on the performance of the numerical solver, whose performance becomes problematic in some cases as reported by \citet{Zhuang2020AdaptiveCA}.
To answer the RQ1, we compare the adjoint method against the full gradient obtained from back-propagation if possible, and if not due to the memory issue, we use the {approximated} gradient obtained from a \emph{surrogate process}.
The surrogate process utilizes the fact that given the total amount of noise, we can denoise the same amount of noise with different numbers of denoising steps~\cite{Song2020DenoisingDI}. Therefore, instead of using the entire denoising steps, we can mimic the original denoising process with fewer function calls, whose gradients can be obtained by back-propagating the forward and denosing process directly.

The gradients obtained from the surrogate process differ from the exact gradients. However, if the accumulated denoising steps can be approximated with fewer denoising steps, we can use the approximated gradients as a proxy of the exact gradients. The surrogate process can also relax the randomness occurring in multiple denoising steps.

\begin{table}[!t]
    \centering
    \begin{tabular}{lccc}
    \toprule
        Defense & Gradient of Def & Robust Accuracy (\%) \\ \midrule
        \multirow{2}{*}{DiffPure~\citep{Nie2022DiffusionMF}} & Adjoint & 74.38±1.03 \\
        & Surrogate & 46.84±1.44 \\ \midrule
        \multirow{2}{*}{GDMP~\citep{Wang2022GuidedDM}} & BPDA & 75.59±1.26 \\
        & Surrogate & 24.06±0.47 \\ \midrule
        \multirow{3}{*}{SODEF~\citep{kang2021stable}} & w/o & 53.69 \\
        & Adjoint & 57.76 \\
        & Full & 49.28 \\ \bottomrule
    \end{tabular}
    \caption{
Robust accuracy of DiffPure, GDMP, and SODEF against attacks ($\ell_\infty(\epsilon = 8/255)$) on CIFAR-10. We use PGD+EOT for DiffPure and GDMP and AutoAttack for SODEF. \textit{Adjoint} calculates full gradients using the adjoint method, and \textit{Surrogate} (or \textit{Full}) calculates approximated (or full) gradients using direct back-propgation. \textit{w/o} is the performance of the underlying classifier without the SODEF.}
    \label{table:RQ1}
\end{table}


\paragraph{RQ1: Is the adjoint method the best way to generate adversarial examples with full gradients?} 
We compare the adjoint method with the full gradient obtained from direct back-propagation of the defense process with the original or surrogate processes.

\autoref{table:RQ1} shows that the robust accuracy of DiffPure \citep{Nie2022DiffusionMF} with the direct back-propagation is 46.84\% on PGD+EOT attack, which is 27.54\% lower than the reported accuracy with the adjoint method. The results show that direct back-propagation is more effective than the adjoint method. Furthermore, we use a surrogate process for GDMP~\citep{Wang2022GuidedDM}, and the robust accuracy is 24.06\%, which is 51.53\% lower than the reported accuracy against the BPDA attack. It can be concluded that, in cases where the gradients of the defense process are unavailable to calculate, the surrogate process can be an alternative to generate adversarial examples.

SODEF \citep{kang2021stable}, a non-diffusion-based purification, originally uses the adjoint method to generate adversarial examples. We evaluate SODEF with direct back-propagation for the attack and observe 49.28\% robust accuracy against AutoAttack, which is lower than 53.69\% of the underlying model without defense. This result suggests that the use of a numerical solver would not be effective for the adversarial attack.

\begin{table}[!t]
    \centering
    \resizebox{\linewidth}{!}{
    \begin{tabular}{cccc}
    \toprule
        \multirow{2}{*}{Threat Model} & \multirow{2}{*}{Defense} & \multirow{2}{*}{Attack} & Robust \\
        & & & Accuracy (\%) \\ \midrule
        \multirow{4}{*}{$\ell_\infty(\epsilon=8/255)$} & \multirow{2}{*}{ADP~\citep{Yoon2021AdversarialPW}} & PGD+EOT & 33.48±0.86 \\
        & & AutoAttack & 59.53±0.87 \\
        & \multirow{2}{*}{DiffPure~\citep{Nie2022DiffusionMF}}  & PGD+EOT & 46.84±1.44 \\
        & & AutoAttack & 63.60±0.81 \\ \midrule
        \multirow{4}{*}{$\ell_2(\epsilon=0.5)$} & \multirow{2}{*}{ADP~\citep{Yoon2021AdversarialPW}} & PGD+EOT & 73.32±0.76 \\
        & & AutoAttack & 79.57±0.38 \\
        & \multirow{2}{*}{DiffPure~\citep{Nie2022DiffusionMF}}  & PGD+EOT & 79.45±1.16 \\
        & & AutoAttack & 81.70±0.84 \\ \bottomrule
    \end{tabular}
    }
    \caption{Robust accuracy of DiffPure and ADP against PGD+EOT and AutoAttack ($\ell_\infty(\epsilon = 8/255)$) on CIFAR-10.}
    \label{table:RQ2}
\end{table}
\paragraph{RQ2: Is AutoAttack still better than the other gradient-based attacks, such as PGD, when the alternative is available?}
AutoAttack has recently been used as a standard method to evaluate defenses due to its robustness against defenses. Although AutoAttack may not be an ideal choice for randomized defenses\footnote{\url{https://github.com/fra31/auto-attack/blob/master/flags_doc.md}}, still many purification methods, such as DiffPure, rely on AutoAttack. However, as shown in \autoref{table:RQ2}, AutoAttack has a lower success rate than PGD+EOT against diffusion-based purification methods. For the $\ell_\infty$ threat model $(\epsilon = 8/255)$, PGD+EOT shows 16.76\% and 26.05\% more attack success rate than AutoAttack against DiffPure and ADP, respectively. We observe a similar result with the $\ell_2$ threat model $(\epsilon = 0.5)$. Therefore, evaluation with PGD+EOT for diffusion-based purification can be useful to evaluate their robustness. Further results of the difference between PGD+EOT against DiffPure with additional settings can be found in \autoref{app:evaluation}.

\begin{table}[!t]
    \centering
    \resizebox{\linewidth}{!}{
        \begin{tabular}{lccc}
        \toprule
            Defense & Type & BPDA & Ours \\ \midrule
            ADP~\citep{Yoon2021AdversarialPW} & DSM+LD & 66.91±1.75 & 33.48±0.86 \\
            DiffPure~\citep{Nie2022DiffusionMF} & Diffusion & 81.45±1.51 & 46.84±1.44 \\
            GDMP~\citep{Wang2022GuidedDM} & Diffusion & 75.59±1.26 & 24.06±0.47 \\
            DISCO~\citep{ho2022disco} & Implicit function & 47.18 & 0.00 \\ \bottomrule
        \end{tabular}
    }
    \caption{Robust accuracy of defenses against BPDA and our full-gradient based attacks ($\ell_\infty(\epsilon = 8/255)$) on CIFAR-10. We report the lowest robust accuracy between PGD and AutoAttack.}
    \label{table:RQ3}
\end{table}
\paragraph{RQ3: Is BPDA still more effective than the best combination of full-gradient attacks?}
BPDA~\citep{obfuscated-gradients} has been widely used to evaluate defenses that can cause gradient obfuscation. Because multiple function calls can cause gradient obfuscation, ADP, GDMP, and DISCO have been evaluated on BPDA as the strongest adaptive white-box attack. However, our evaluation shows that BPDA has a lower attack success rate than the attacks using direct gradients of the defense process, as shown in \autoref{table:RQ3}. Against PGD+EOT using direct gradients of defense process, ADP and GDMP show robust accuracy of 33.48\% and 24.06\%, respectively, significantly lower than the reported accuracy with BPDA~\citep{Yoon2021AdversarialPW, Wang2022GuidedDM}. DISCO even has 0\% robust accuracy. From the results, we suggest that the direct gradients of the defense process need to be tested to check the robustness.

\paragraph{Recommendation.}
We propose an overall guideline for evaluating diffusion-based purifications as follows. We recommend using PGD+EOT rather than AutoAttack. When calculating gradients, it is best to directly back-propagate the full defense process. If this is unavailable due to memory constraints, using the surrogate process rather than the adjoint method is recommended. 
Note that our recommendation generally follows the suggestions made by \citet{croce2022evaluating} but is more tailored for the diffusion-based purification.

\section{Analysis of Hyperparameters}
\label{sec:hyperparameter}

The performance of diffusion-based purification is significantly influenced by varying hyperparameter configurations. In this section, we explore the importance of hyperparameters in defense processes.

\subsection{Experimental Settings}
\label{sec:param_settings}

Understanding the importance of hyperparameters can help build a better defense mechanism. We investigate the effect of various hyperparameters of diffusion-based purification methods to determine the most robust configuration for the adaptive attack. 
Specifically, the following three hyperparameters are evaluated 1) the number of forward steps, 2) the number of denoising steps, and 3) the number of purification steps. In addition, we re-evaluate the efficiency of several techniques proposed in previous works under our defense scheme.

We evaluate the purification against PGD+EOT on CIFAR-10. We provide the additional results on CIFAR-10 and ImageNet in \autoref{app:hyperparameter}. Although we do not report ImageNet results in the main text, the overall findings are similar to those from CIFAR-10. We use a naturally pretrained WideResNet-28-10~\citep{Zagoruyko2016WideRN} as an underlying classifier provided by Robustbench~\citep{croce2021robustbench}. For a diffusion model, we use pretrained DDPM++~\citep{Song2020ScoreBasedGM}. The variances for the diffusion model are linearly increasing from $\beta_1 = 10^{-4}$ to $\beta_T = 0.02$ when $T = 1000$ \citep{Ho2020DenoisingDP}. We use two different denoising models: DDPM~\citep{Ho2020DenoisingDP} and DDIM~\citep{Song2020DenoisingDI}. 

For all experiments, we report the mean and standard deviation over five runs to measure the standard and robust accuracy. PGD uses 200 update iterations. 20 samples are used to compute EOT. Following the settings in DiffPure~\citep{Nie2022DiffusionMF}, we use a fixed subset of 512 randomly sampled images. To calculate gradients, we use direct gradients of the entire process. If impossible, we compute the approximated gradients from a surrogate process. In each experiment, we explain the defense process and the surrogate process in more detail.

\begin{figure}[t!]
    \centering
    \includegraphics[width=0.9\linewidth]{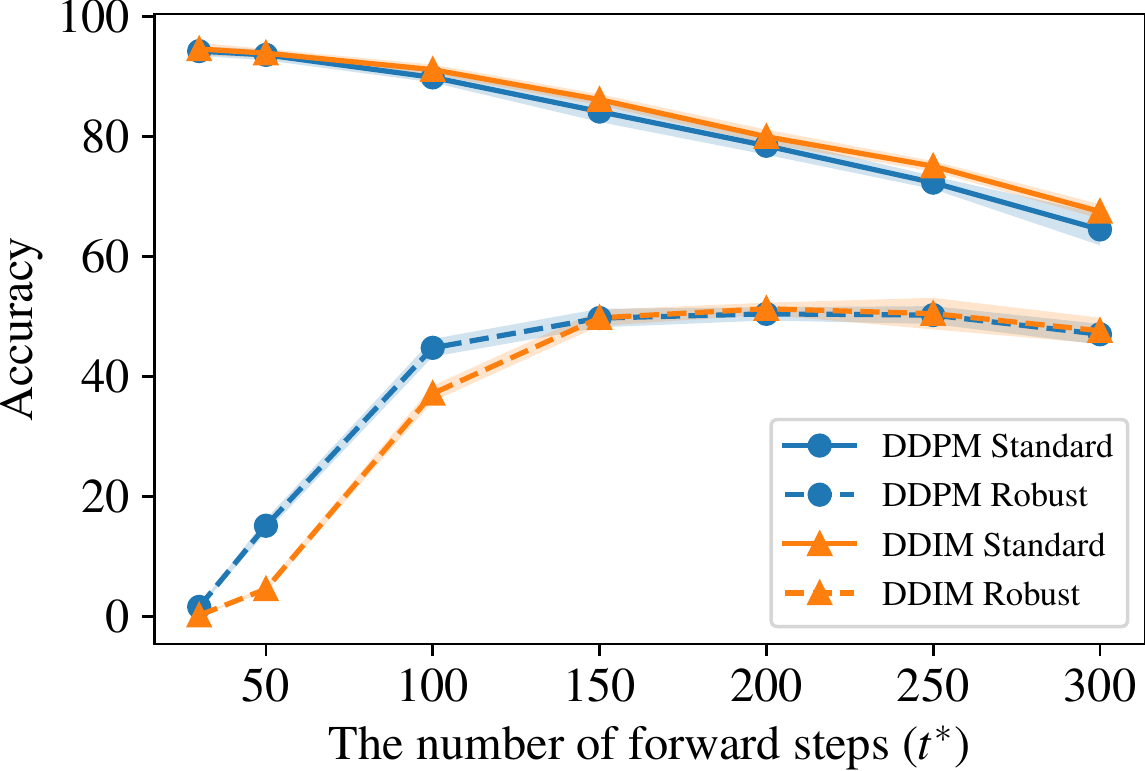}
    \caption{Standard and robust accuracy as we change the number of forward steps against PGD+EOT $\ell_\infty (\epsilon = 8/255)$ on CIFAR-10. Five denoising steps for both attack and defense are used.}
    \label{fig:forward_steps}
\end{figure}
\subsection{The Number of Forward Steps}
\label{sec:forward_steps}
We explore the effect of forward noising steps on robustness by varying the number of forward steps from 30 to 300, resulting in the changes of total variance ranged from 0.012 to 0.606. The same number of forward steps are used for both attack and defense, and we set five denoising steps for attack and defense for all experiments.

As shown in \autoref{fig:forward_steps}, the standard accuracy continuously decreases as the number of forward steps increases since more forward steps induce more noise.
The robust accuracy increases first and decreases after 200 forward steps, i.e., $t^* = 200$. When the number of forward steps is small, the DDPM is more robust than the DDIM. However, DDIM shows better accuracy for both standard and robust than DDPM after 200 forward steps.

\begin{figure}[t!]
    \centering
    \includegraphics[width=0.9\linewidth]{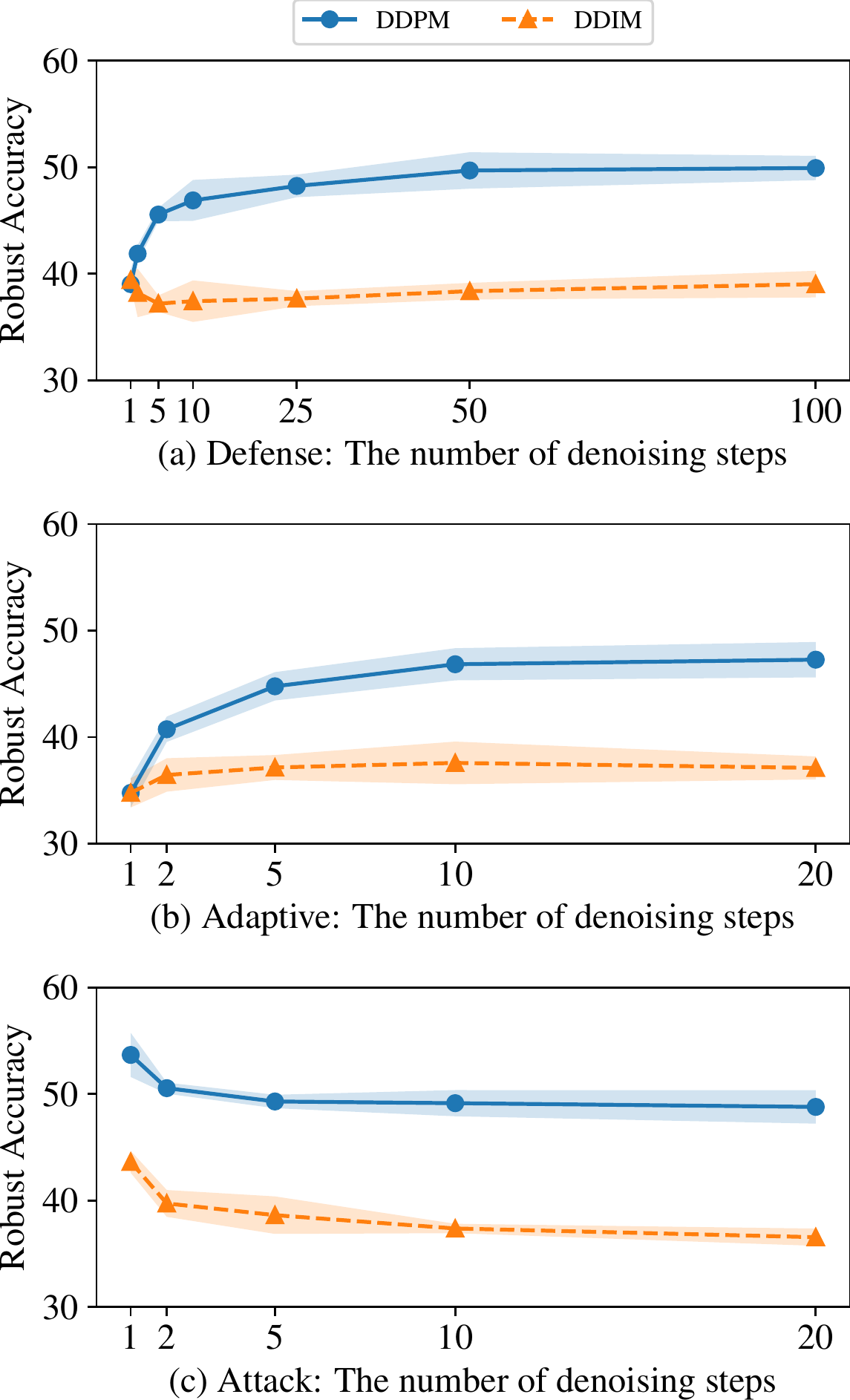}
    \caption{Robust accuracy as we change the number of denoising steps against PGD+EOT $\ell_\infty (\epsilon = 8/255)$ on CIFAR-10. We change the number of denoising steps in (a) defense, (b) both, and (c) attack for each experiment with the other hyperparameters fixed.}
    \label{fig:sampling_steps_t100}
\end{figure}

\subsection{The Number of Denoising Steps}
\label{sec:denoising_steps}
Defenders may use fewer denoising steps to accelerate the defense process. From the other perspective, attackers may want to use fewer denoising steps than those used in the defense due to memory constraints. We explore the influence of the number of denoising steps  through the following three experimental settings:
\begin{itemize}
    \item[(a)] The number of denoising steps in attack is set to five, and the number of denoising steps in defense is ranged from one to the maximum number of denoising steps.
    \item[(b)] The number of denoising steps in both the attack and defense are the same, ranging from one to 20.
    \item[(c)] The number of denoising steps in defense is set to the maximum number of denoising steps, and the number of denoising steps in attack is ranged from one to 20.\footnote{The 20 denoising steps is the maximum limit of 40GB of memory.}
\end{itemize}

The results are displayed in \autoref{fig:sampling_steps_t100}.
From the defense perspective, the results of (a) and (b) demonstrate that more denoising steps can improve robustness. DDPM gains more advantage from having more denoising steps than DDIM.
(c) shows the effect of the number of denoising steps in the attack. As the number of denoising steps increases, the attack success rate slightly increases. However, we also find that increasing the number of denoising steps in an attack can decrease the attack success rate when the number of forward steps is 200 (i.e., $t^* = 200$).

\begin{figure}[t!]
    \centering
    \includegraphics[width=0.9\linewidth]{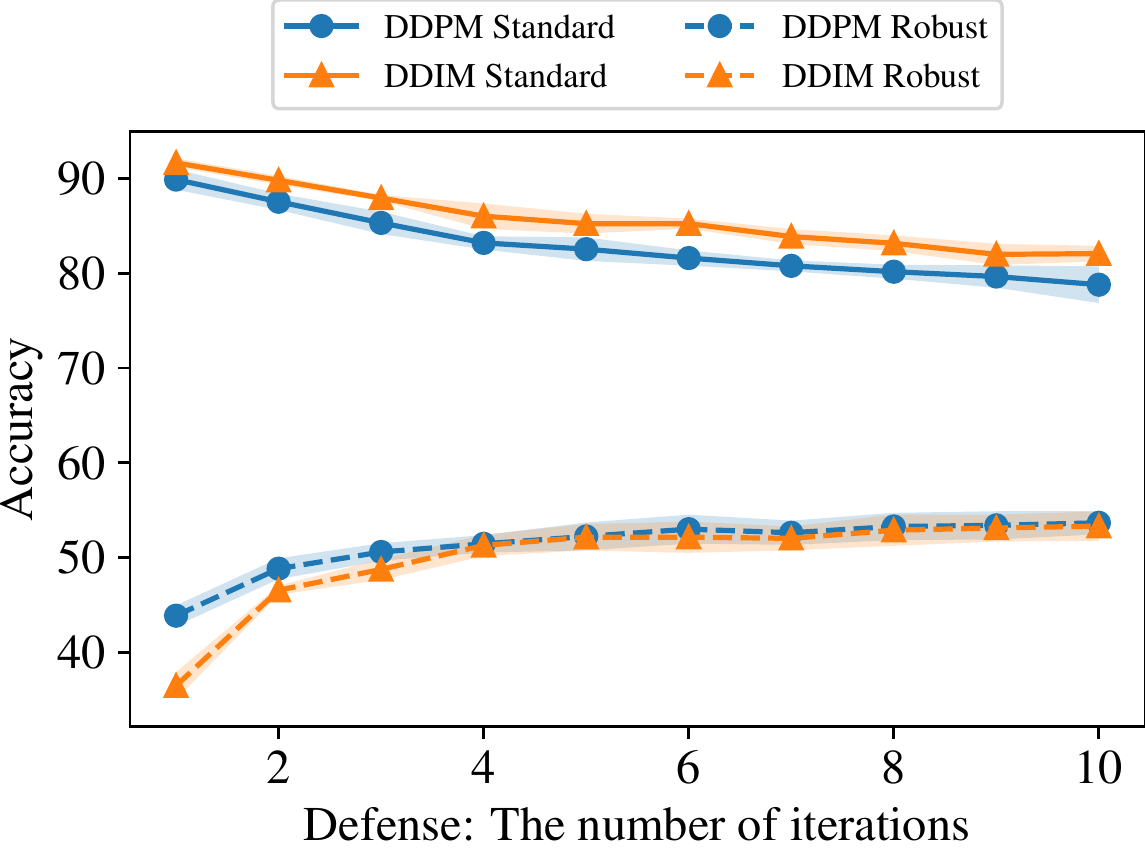}
    \caption{
    The number of purification steps in defense and its influence to the standard and robust accuracy against PGD+EOT $\ell_\infty (\epsilon = 8/255)$ on CIFAR-10. The number of forward steps is 100 (i.e., $t^* = 100$). The reported robust accuracy is the lowest performance of all settings of the number of purification steps of the attack.}
    \label{fig:defense_num_iterations_t100}
\end{figure}
\begin{figure}[t!]
    \centering
    \includegraphics[width=0.9\linewidth]{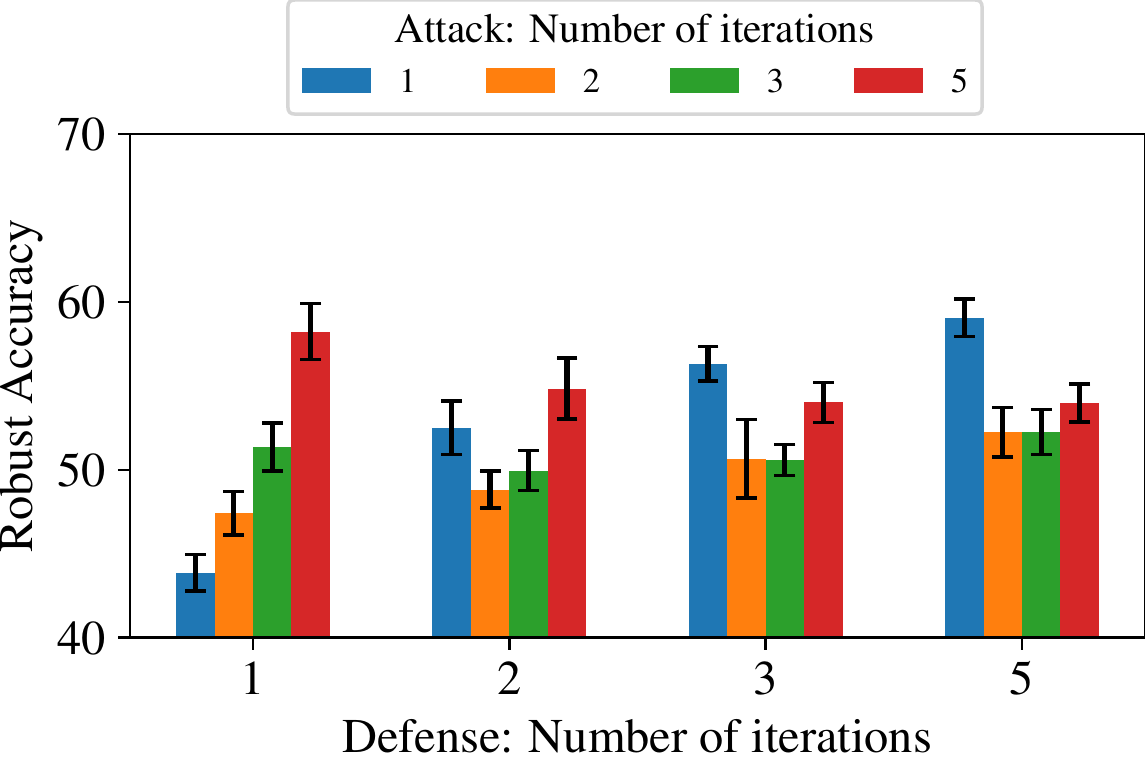}
    \caption{
    The number of purification steps during attacks and its influence  on the robust accuracy against PGD+EOT $\ell_\infty (\epsilon = 8/255)$ with CIFAR-10. The number of forward steps is fixed as 100 (i.e., $t^* = 100$).}
    \label{fig:attack_num_iterations_t100}
\end{figure}
\subsection{The Number of Purification Steps}
\label{sec:purification_steps}
Although a single forward and reverse process can purify the input image, one can apply the purification process multiple times as proposed in \citet{Wang2022GuidedDM}. We denote the number of forward and denoising processes as the number of \textit{purification step}. Similar to the case of the denoising step, computing the gradients of multiple purification steps is impossible due to memory constraints in most cases. 

The number of purification steps can also differ between attack and defense. Through experiments, we measure the changes in robust accuracy with the different number of purification steps in the defense and attack.
For all experiments, we fixed the number of forward steps to 100 ($t^* = 100$), and the number of denoising steps is set to five.

\autoref{fig:defense_num_iterations_t100} shows the standard and robust accuracy with a varying number of purification steps in defense. The robust accuracy increases as the number of purification steps increases while the standard accuracy steadily decreases. 
\autoref{fig:attack_num_iterations_t100} shows the effect of the number of purification steps in the attack. When the number of purification steps in defense is one or two, the same number of purification steps in attack is the most effective. However, as we set the number of purification steps in defense to three and five, two and three purification steps in attack show a better attack success, respectively.

\subsection{Other Techniques}
\label{sec:other_tech}

We evaluate several other techniques proposed in earlier work~\citep{Yoon2021AdversarialPW, Nie2022DiffusionMF, Wang2022GuidedDM} within our new evaluation framework.


\begin{table}[t!]
    \centering
    \begin{tabular}{lccc}
    \toprule
        \multirow{2}{*}{Guidance} & \multicolumn{3}{c}{Accuracy (\%)} \\
        ~ & Standard & BPDA & PGD+EOT \\ \midrule
        No guide & 87.70±0.46 & 75.23±0.61 & 38.44±0.59 \\
        MSE & 89.96±0.40 & 75.59±1.26 & 24.06±0.47 \\
        SSIM & 93.75±0.39 & 74.02±1.17 & 6.88±0.21 \\ \bottomrule
    \end{tabular}
    \caption{Standard and robust accuracy of GDMP~\citep{Wang2022GuidedDM} against BPDA and PGD+EOT $\ell_\infty (\epsilon = 8/255)$ on CIFAR-10. We compare two types of guidance, MSE and SSIM, and the defense without guidance.}
    \label{table:guidance_evaluation}
\end{table}
\begin{figure}[t!]
    \centering
    \includegraphics[width=0.9\linewidth]{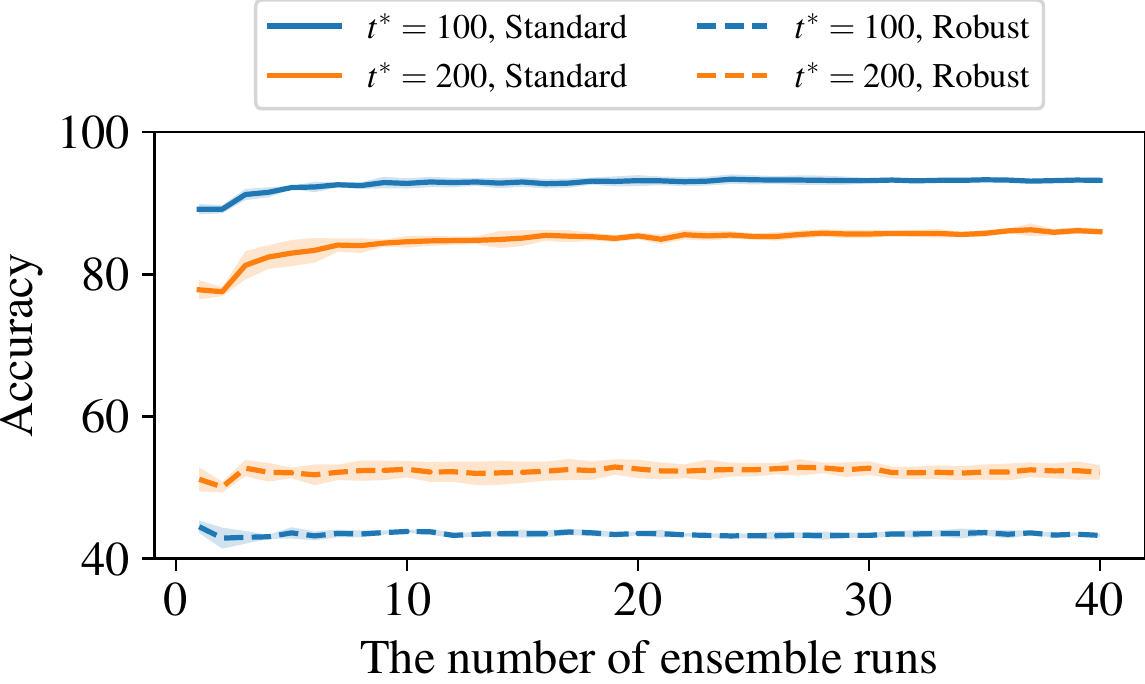}
    \caption{Standard and robust accuracy against PGD+EOT $\ell_\infty (\epsilon = 8/255)$ on CIFAR-10 when using an ensemble with a different number of purification runs. Five denoising steps are used for the surrogate process of the attack.}
    \label{fig:ensemble}
\end{figure}

\paragraph{Guidance.}
GDMP~\citep{Wang2022GuidedDM} proposes to use gradients of a distance between an original input example and a target example to preserve semantic information while denoising. They show guidance can improve robustness against preprocessor-blind attacks. However, as shown in \autoref{table:guidance_evaluation}, when the gradients of the surrogate process are used in the attack, the guidance of GDMP decreases the robust accuracy. Specifically, the defense with guidance using the SSIM similarity has 6.88\% robust accuracy, which is 31.56\% lower than the defense without guidance.

\paragraph{Ensemble of multiple purification runs.}
ADP~\citep{Yoon2021AdversarialPW} uses the ensemble of multiple purification runs as the predicted label to mitigate the randomness in the defense. For diffusion-based purification methods, as shown in \autoref{fig:ensemble}, multiple purification runs especially can help improve standard accuracy while the robust accuracy keeps the same level. In particular, for $t^*=200$ with 40 purification runs, standard accuracy is 8.17\% higher than the case without ensemble.

\begin{table}[t!]
    \centering
    \begin{tabular}{lcc}
    \toprule
        Underlying Classifier & $t^*$ & Robust Accuracy (\%) \\ \midrule
        \multirow{3}{*}{TRADES~\citep{zhang2019theoretically}} & 0 & 55.32 \\
        & 100 & 54.02±0.98 \\
        & 200 & 51.52±1.96 \\
        \midrule
        \multirow{3}{*}{\citet{Gowal2021ImprovingRU}} & 0 & 69.03 \\
        & 100 & 58.24±0.49 \\
        & 200 & 52.97±1.38 \\ \bottomrule
    \end{tabular}
    \caption{Combination of diffusion models with adversarial training evaluated on CIFAR-10 against PGD+EOT $\ell_\infty (\epsilon = 8/255)$. Five denoising steps are used for both attack and defense.}
    \label{table:combi_with_adv}
\end{table}
\begin{table}[!t]
    \centering
    \begin{tabular}{cccc}
    \toprule
        $t^*$ & Defense & Attack & Robust Accuracy (\%) \\ \midrule
        \multirow{4}{*}{100} & \multirow{2}{*}{DDPM} & DDPM & 44.77±1.48 \\
        & & DDIM & 46.68±1.25 \\
        & \multirow{2}{*}{DDIM} & DDPM & 40.51±1.01 \\
        & & DDIM & 37.15±1.31 \\ \midrule
        \multirow{4}{*}{200} & \multirow{2}{*}{DDPM} & DDPM & 50.43±1.11 \\
        & & DDIM & 52.15±1.88 \\
        & \multirow{2}{*}{DDIM} & DDPM & 53.63±1.11 \\
        & & DDIM & 51.29±1.00 \\ \bottomrule
    \end{tabular}
    \caption{Robust accuracy against PGD+EOT $\ell_\infty (\epsilon = 8/255)$ on CIFAR-10 when using a denoising model in attack different from the denoising model in defense.}
    \label{table:sampler}
\end{table}

\paragraph{Combination with adversarial training.}
An adjoint method based DiffPure~\citep{Nie2022DiffusionMF} shows robustness can be improved by using diffusion models together with adversarial training. However, as shown in \autoref{table:combi_with_adv}, the adversarial training with purification shows lower robustness than the classifier without purification.

\begin{table*}[!t]
    \centering
    \begin{subtable}{.49\linewidth}
        \resizebox{\linewidth}{!}{
        \begin{tabular}{cclccc}
        \toprule
            & Type & Method & Standard & PGD & AutoAttack \\ \midrule 
            \multirow{6}{*}{\rotatebox[origin=c]{90}{WRN-28-10}} & \multirow{3}{*}{AT} & \citet{Gowal2021ImprovingRU} & 87.51 & 66.01 & 63.38 \\
            & & \citet{Gowal2020UncoveringTL}* & 88.54 & 65.93 & 62.76 \\
            & & \citet{Pang2022RobustnessAA} & 88.62 & 64.95 & 61.04 \\ \cmidrule{2-6}
            & \multirow{3}{*}{AP} & \citet{Yoon2021AdversarialPW} & 85.66±0.51 & 33.48±0.86 & 59.53±0.87 \\
            & & \citet{Nie2022DiffusionMF} & 90.07±0.97 & 46.84±1.44 & 63.60±0.81 \\
            & & Ours & 90.16±0.64 & 55.82±0.59 & 70.47±1.53 \\ \midrule

            \multirow{6}{*}{\rotatebox[origin=c]{90}{WRN-70-16}} & \multirow{3}{*}{AT} & \citet{Rebuffi2021FixingDA}* & 92.22 & 69.97 & 66.56 \\
            & & \citet{Gowal2021ImprovingRU} & 88.75 & 69.03 & 66.10 \\
            & & \citet{Gowal2020UncoveringTL}* & 91.10 & 68.66 & 65.87 \\ \cmidrule{2-6}
            & \multirow{3}{*}{AP} & \citet{Yoon2021AdversarialPW} & 86.76±1.15 & 37.11±1.35 & 60.86±0.56 \\
            & & \citet{Nie2022DiffusionMF} & 90.43±0.60 & 51.13±0.87 & 66.06±1.17 \\
            & & Ours & 90.53±0.14 & 56.88±1.06 & 70.31±0.62 \\ \bottomrule
        \end{tabular}
        }
    \end{subtable}
    \begin{subtable}{.49\linewidth}
        \resizebox{\linewidth}{!}{
        \begin{tabular}{cclccc}
        \toprule
            & Type & Method & Standard & PGD & AutoAttack \\ \midrule
            \multirow{6}{*}{\rotatebox[origin=c]{90}{WRN-28-10}} & \multirow{3}{*}{AT} & \citet{Rebuffi2021FixingDA}* & 91.79 & 85.05 & 78.80 \\
            & & \citet{Augustin2020AdversarialRO}$^\dagger$ & 93.96 & 86.14 & 78.79 \\
            & & \citet{Sehwag2021RobustLM}$^\dagger$ & 90.93 & 83.75 & 77.24 \\ \cmidrule{2-6}
            & \multirow{3}{*}{AP} & \citet{Yoon2021AdversarialPW} & 85.66±0.51 & 73.32±0.76 & 79.57±0.38 \\
            & & \citet{Nie2022DiffusionMF} & 91.41±1.00 & 79.45±1.16 & 81.70±0.84 \\
            & & Ours & 90.16±0.64 & 83.59±0.88 & 86.48±0.38 \\ \midrule

            \multirow{6}{*}{\rotatebox[origin=c]{90}{WRN-70-16}} & \multirow{3}{*}{AT} & \citet{Rebuffi2021FixingDA}* & 95.74 & 89.62 & 82.32 \\
            & & \citet{Gowal2020UncoveringTL}* & 94.74 & 88.18 & 80.53 \\
            & & \citet{Rebuffi2021FixingDA} & 92.41 & 86.24 & 80.42 \\ \cmidrule{2-6}
            & \multirow{3}{*}{AP} & \citet{Yoon2021AdversarialPW} & 86.76±1.15 & 75.66±1.29 & 80.43±0.42 \\
            & & \citet{Nie2022DiffusionMF} & 92.15±0.72 & 82.97±1.38 & 83.06±1.27 \\
            & & Ours & 90.53±0.14 & 83.75±0.99 & 85.59±0.61 \\ \bottomrule
        \end{tabular}
        }
    \end{subtable}
    \caption{Standard and robust accuracy against PGD+EOT (left: $\ell_\infty (\epsilon = 8/255)$, right: $\ell_2 (\epsilon = 0.5)$) on CIFAR-10. Adversarial Training (AT) and Adversarial Purification (AP) methods are evaluated. $^\dagger$ This method uses WideResNet-34-10 as a classifier. * This method is trained with extra data.}
    \label{table:best_cifar10}
\end{table*} 

\paragraph{Transferability of gradients from different samplers in the attack.}
One may employ a sampler of diffusion models to generate adversarial examples different from those used in defense. For example, attacks using gradients from DDPM could be transferred to the defense using DDIM and vice-versa. We test whether the gradients from a different sampler of denoising models can improve the attack success rate. As shown in \autoref{table:sampler}, although the transferred attack is valid, the attack success rates using different samplers are slightly lower than those using the original samplers.

\begin{table}[!t]
    \centering
    \resizebox{\linewidth}{!}{
        \begin{tabular}{clcc}
        \toprule
            \multirow{2}{*}{Type} & \multirow{2}{*}{Method} & \multicolumn{2}{c}{Accuracy (\%)} \\
            ~ & ~ & Standard & Robust \\ \midrule
            \multirow{3}{*}{AT} & \citet{Salman2020DoAR} & 63.86 & 39.11 \\
            & \citet{robustness} & 62.42 & 33.20 \\
            & \citet{Wong2020FastIB} & 53.83 & 28.04 \\ \midrule
            \multirow{2}{*}{AP} & \citet{Nie2022DiffusionMF} & 71.48±0.66 & 38.71±0.96 \\
            & Ours & 70.74±0.91 & 42.15±0.64 \\ \bottomrule
        \end{tabular}
    }
    \caption{Standard and robust accuracy against PGD+EOT $\ell_\infty (\epsilon = 4/255)$ on ImageNet. ResNet-50 is used as a classifier.}
    \label{table:linf_imagenet_resnet50}
\end{table}

\section{Gradual Noise-Scheduling for Multi-Step Purification}
\label{sec:gradual_sampling}

In this section, we propose a new sampling strategy for diffusion-based purification and compare the performance with other state-of-the-art defenses.

\textbf{Gradual noise-scheduling strategy.} 
As highlighted in \autoref{sec:hyperparameter}, selecting appropriate hyperparameter values is essential to improve robustness. Thus, we conduct an extensive exploration of hyperparameter settings to maximize robust accuracy. In particular, we mainly focus on the fact that each purification step can contain a different number of forward steps. We empirically find that fewer forward steps in the first few purification steps can improve robustness.

Based on this observation, for CIFAR-10, we set the number of forward steps as $\{30 \times 4, 50 \times 2, 125 \times 2\}$ for eight purification steps. For ImageNet and SVHN, we set the number of forward steps as $\{30 \times 4, 50 \times 2, 200 \times 2\}$ and $\{30 \times 4, 50 \times 2, 80 \times 2\}$, respectively. We set the number of denoising steps to equal the number of forward steps for all purification steps. We use the DDPM and an ensemble of ten purification runs.

\begin{table}[!t]
    \centering
    \resizebox{\linewidth}{!}{
        \begin{tabular}{cccc}
            \toprule
            \multirow{2}{*}{Type} & \multirow{2}{*}{Defense} & \multicolumn{2}{c}{Accuracy (\%)} \\
            & & Standard & Robust \\ \midrule
            \multirow{3}{*}{AT} & \citet{Rade2022ReducingEM} & 93.08 & 52.83 \\
            & \citet{Gowal2020UncoveringTL} & 92.87 & 56.83 \\
            & \citet{Gowal2021ImprovingRU} & 94.15 & 60.90 \\ \midrule
            \multirow{2}{*}{AP} & \citet{Nie2022DiffusionMF} & 97.85±0.53 & 34.30±0.41 \\
            & Ours & 95.55±0.40 & 49.65±1.06 \\ \bottomrule
        \end{tabular}
    }
    \caption{Standard and robust accuracy against attacks $\ell_\infty (\epsilon = 8/255)$ on SVHN. Adversarial training methods are evaluated on AutoAttack, and adversarial purification methods are evaluated on PGD+EOT. WideResNet-28-10 is used as a classifier except for \citet{Rade2022ReducingEM}, which uses ResNet-18.}
    \label{table:best_svhn}
\end{table}

\textbf{Experimental settings.} 
We conduct evaluations on three datasets, CIFAR-10~\citep{krizhevsky2009learning}, ImageNet~\cite{deng2009imagenet}, and SVHN~\citep{Netzer2011ReadingDI}. We use three diffusion model architectures, DDPM++~\citep{Song2020ScoreBasedGM}, Guided Diffusion~\citep{Dhariwal2021DiffusionMB}, and DDPM~\citep{Ho2020DenoisingDP} for each dataset. We use pretrained models for CIFAR-10 and ImageNet, but we trained a model for SVHN. Pretrained WideResNet-28-10, WideResNet-70-16, and ResNet-50~\citep{Zagoruyko2016WideRN, He2015DeepRL} are served as baseline classifiers. We compare our method with adversarial training and diffusion-based purification methods. 
We evaluate diffusion-based purification methods on the PGD+EOT attack with 200 update iterations, except for ImageNet, which uses 20 iterations. We set the number of EOT to 20. For adversarial training methods, we use 20 update iterations for the PGD attack.
For our method, we report the worst robust accuracy with surrogate processes. We explain the detailed settings in \autoref{app:gradual_sampling}.

\textbf{Results.}
\autoref{table:best_cifar10} shows the defense performance against $\ell_\infty (\epsilon = 8/255)$ and $\ell_2 (\epsilon = 0.5)$ threat models on CIFAR-10, respectively.
Our method outperforms other diffusion-based purification methods. Specifically, compared to DiffPure on $\ell_\infty$ PGD attack, our method improves robust accuracy by 8.98\% with WideResNet-28-10 and by 5.75\% with WideResNet-70-16, respectively. Despite the improvement in robustness, the purification methods perform worse than the adversarial training methods.
\autoref{table:linf_imagenet_resnet50} shows the performance against $\ell_\infty (\epsilon = 4/255)$ threat model on ImageNet. Our method outperforms both adversarial training and purification methods. Compared to DiffPure and \citet{Salman2020DoAR}, our method improves robust accuracy by 4.34\% and 3.94\%, respectively. 
Results on SVHN against threats model $\ell_\infty (\epsilon = 8/255)$ are similar with CIFAR-10. Although our method improves robust accuracy by 15.35\% compared to the DiffPure framework, which uses $t^* = 0.075$, our method performs worse than the adversarial training methods.

We additionally compare the robustness of our defense strategy with other adversarial purification methods against BPDA ($\ell_\infty (\epsilon = 8/255)$). As shown in \autoref{table:best_BPDA}, our proposed method outperforms all other adversarial purification methods, achieving a robust accuracy of 88.40\%, 6.95\% greater than the robust accuracy of DiffPure. Furthermore, \autoref{table:best_other_attack} shows our robustness against other attacks, including the Square attack~\citep{Andriushchenko2019SquareAA}, a black-box attack. Our defense shows strong robustness higher than 80\% against all attacks.

\begin{table}[!t]
    \centering
    \resizebox{\linewidth}{!}{
        \begin{tabular}{lccc}
        \toprule
            \multirow{2}{*}{Method} & \multirow{2}{*}{Purification} & \multicolumn{2}{c}{Accuracy (\%)} \\
            ~ & ~ & Standard & Robust \\ \midrule
            \citet{Song2017PixelDefendLG} & Gibbs Update & 95.00 & 9.00 \\
            \citet{Yang2019MENetTE} & Mask+Recon & 94.00 & 15.00 \\
            \citet{Hill2020StochasticSA} & EBM+LD & 84.12 & 54.90 \\ \midrule
            \citet{Yoon2021AdversarialPW} & DSM+LD & 85.66±0.51 & 66.91±1.75 \\
            \citet{Nie2022DiffusionMF} & Diffusion & 90.07±0.97 & 81.45±1.51 \\ 
            Ours & Diffusion & 90.16±0.64 & 88.40±0.88 \\ \bottomrule
        \end{tabular}
    }
    \caption{Standard and robust accuracy against BPDA+EOT $\ell_\infty (\epsilon = 8/255)$ on CIFAR-10. WideResNet-28-10 is used as the underlying classifier architecture.}
    \label{table:best_BPDA}
\end{table}
\begin{table}[!t]
    \centering
    \begin{tabular}{lc}
    \toprule
        Attack & Robust Accuracy (\%) \\ \midrule
        Square~\citep{Andriushchenko2019SquareAA} & 89.38±0.26 \\
        FAB~\citep{Croce2019MinimallyDA} & 89.18±0.60 \\
        Deep Fool~\citep{MoosaviDezfooli2015DeepFoolAS} & 82.32±0.14 \\
        FMN Attack~\citep{Pintor2021FastMA} & 80.86±1.80 \\ \bottomrule
    \end{tabular}
    \caption{Robust accuracy of our defense strategy against several threat models $\ell_\infty (\epsilon = 8/255)$ on CIFAR-10. Square attack is a black-box attack, and the others are white-box attacks.}
    \label{table:best_other_attack}
\end{table}

\section{Related Work}
\label{Related Work}

Adversarial training~\citep{Madry2017TowardsDL,zhang2019theoretically} is one of the most successful adversarial defense methods. These methods train a classifier with adversarial examples in a training phase. \citet{zhang2019theoretically} and \citet{Pang2022RobustnessAA} propose loss functions that can effectively utilize the trade-off between robustness and accuracy. \citet{Huang2021ExploringAI} analyze architectural factors with respect to robustness. \citet{Rebuffi2021DataAC} and \citet{Gowal2021ImprovingRU} improve robustness by utilizing data augmentations.

Adaptive test-time defenses purify adversarial examples using extra neural networks that utilize techniques from other domains. ADP~\citep{Yoon2021AdversarialPW} jointly uses denoising score matching and Langevin dynamics for purification. DiffPure~\citep{Nie2022DiffusionMF} demonstrates from the stochastic differential equations (SDE) perspective that diffusion models can purify adversarial examples. GDMP~\citep{Wang2022GuidedDM} uses the guidance of diffusion models to recover adversarial examples as similar as possible to the original examples. SODEF~\citep{kang2021stable} uses a Lyapunov-stable ODE block so that the input converges to a stable point that can be correctly classified. DISCO~\citep{ho2022disco} is one of the denoising models that predict clean RGB value using local implicit functions.

\section{Conclusion}
Throughout the paper, we first analyze the current evaluation methods for diffusion-based adversarial purification and then propose a recommendation for the reliable evaluation of the robustness of adversarial purification. We further investigate the influence of hyperparameters of the diffusion model on the robustness of the purification. Based on our analysis, we propose a new strategy to maximize the benefit of the purification methods.

\paragraph{Acknowledgements} This work was partly supported by Institute of Information \& communications Technology Planning \& Evaluation (IITP) grant funded by the Korea government (MSIT) (No.2019-0-01906, Artificial Intelligence Graduate School Program (POSTECH)) and National Research Foundation of Korea(NRF) grant funded by the Korea government(MSIT) (No. RS-2023-00217286) and National Research Foundation of Korea (NRF) grant funded by the Korea government (MSIT) (NRF-2021R1C1C1011375)

{\small
\bibliographystyle{plainnat}
\bibliography{egbib}
}

\clearpage
\appendix
\onecolumn

\section{Defenses and Evaluation Configurations}
\label{app:defense_and_evaluation}

\subsection{Attack Configurations}
\label{app:attack_configuration}
We use PGD, BPDA, and AutoAttack for the evaluation on CIFAR-10. PGD uses 200 update iterations, and BPDA and AutoAttack use 100 update iterations. All the attacks use 20 EOT samples. The step size of PGD and BPDA is 0.007. For randomized defenses, such as DiffPure~\citep{Nie2022DiffusionMF}, we use the random version of AutoAttack, and for static defenses, such as SODEF~\citep{kang2021stable} and DISCO~\citep{ho2022disco}, we use the standard version. For diffusion-based purification methods, following the settings in DiffPure, we use a fixed subset of 512 randomly sampled images for all experiments.

\subsection{Diffusion-Based Purification}
\label{app:diffusion_based_purification}
Diffusion-based purification methods follow the algorithm proposed by DiffPure~\citep{Nie2022DiffusionMF}. Diffusion-based purification partialy utilizes the forward and denoising processes. Algorithm~\ref{alg:forward_process} displays the complete forward process of diffusion-based purification. Using the ``notable property" of the forward process~\citep{Ho2020DenoisingDP}, we can sample $\mathbf{x}_{t^*}$ in a single step:
\begin{equation}
    q(\mathbf{x}_{t^*} | \mathbf{x}_{0}) = \mathcal{N}(\mathbf{x}_{t^*}; \sqrt{\alpha_{t^*}} \mathbf{x}_{0}, (1 - \alpha_{t^*})\textbf{I}),
\end{equation}
where $\alpha_{t^*} := \prod_{i = 1}^{t^*}(1 - \beta_i)$. 
Algorithm~\ref{alg:denoising_process} displays the complete denoising process of diffusion-based purification. \citet{Song2020DenoisingDI} find that denoising process can be accelerated by a linearly increasing sub-sequence $\{\tau_0, \dotsc, \tau_s\}$ of $[0, \ldots, t^*]$ with $\tau_0 = 0$ and $\tau_s = t^*$ where $s \leq t^*$. Throughout all experiments, we only consider sub-sequences having uniform step size. Given $\sigma_{\tau_i}(\eta) = \eta\sqrt{(1 - \alpha_{\tau_{i-1}})/(1 - \alpha_{\tau_i})}\sqrt{1 - \alpha_{\tau_i}/\alpha_{\tau_{i-1}}}$ for all timesteps, the denoising process is DDPM when $\eta = 1$ and DDIM when $\eta = 0$.

\begin{algorithm}[H]
    \centering
    \caption{Forward process of diffusion-based purification}\label{alg:forward_process}
    \begin{algorithmic}[1]
        \State \textbf{Input}: image $\mathbf{x}_0$, maximum timestep $t^*$
        \State $\boldsymbol{\epsilon} \sim \mathcal{N}(\boldsymbol{0}, \mathbf{I})$
        \State $\mathbf{x}_{t^*} = \sqrt{\alpha_{t^*}}\mathbf{x}_0 + \sqrt{1 - \alpha_{t^*}}\boldsymbol{\epsilon}$
        \State \textbf{Return} $\mathbf{x}_{t^*}$
    \end{algorithmic}
\end{algorithm}
\begin{algorithm}[H]
    \centering
    \caption{Denoising process of diffusion-based purification}\label{alg:denoising_process}
    \begin{algorithmic}[1]
        \State \textbf{Input}: noisy image $\mathbf{x}_{t^*}$, timestep schedule $\{\tau_0, \tau_1, \dotsc, \tau_s\}$
        \For{$i = s, \dotsc, 1$}
        \State $\boldsymbol{\epsilon} \sim \mathcal{N}(\boldsymbol{0}, \mathbf{I})$
        \State $\mathbf{x}_{\tau_{i-1}} = \sqrt{\alpha_{\tau_{i-1}}}\bigg(
        {\mathbf{x}_{\tau_{i}} - \sqrt{1 - \alpha_{\tau_{i}}}\mathbf{\epsilon}_\theta (\mathbf{x}_{\tau_{i}}) \over \sqrt{\alpha_{\tau_{i}}}}\bigg) + \sqrt{1 - \alpha_{\tau_{i-1}} - \sigma^2_{\tau_{i}}} \cdot \epsilon_\theta (\mathbf{x}_{\tau_{i}}) + \sigma_{\tau_{i}} \boldsymbol{\epsilon}$
        \EndFor
        \State \textbf{Return}  $\mathbf{x}_{0}$
    \end{algorithmic}
\end{algorithm}

\subsection{ADP~\citep{Yoon2021AdversarialPW}}
ADP uses a score-based model trained with denoising score matching:
\begin{equation}
\mathbb{E}_t\mathbb{E}_{q(\mathbf{x}_t|\mathbf{x})p_{\text{data}}(\mathbf{x})} \Bigg[\frac{1}{2}\| \mathbf{s}_\theta(\mathbf{x}_t) - \nabla_{\mathbf{x}_t} \log q(\mathbf{x}_t|\mathbf{x}) \|^2 \Bigg],
\end{equation}
where $p_{\text{data}}$ is a data distribution, $t$ is a scale of perturbation, and $\mathbf{s}_\theta$ is a score-based model. To clean an adversarial example, ADP uses a deterministic version of Langevin dynamics:
\begin{equation}
\mathbf{x}_t = \mathbf{x}_{t - 1} + \alpha_{t - 1} \mathbf{s}_\theta(\mathbf{x}_{t - 1}),
\end{equation}
where $\alpha_{t-1}$ is a step size. Before its purification process, ADP adds Gaussian noise to the input, which can improve robustness. In its original implementation, ADP is evaluated on BPDA. However, since ADP uses up to eight steps for purification, calculating the full gradients of the defense is possible. Therefore, we use the full gradients to generate adversarial examples in our evaluation.

\subsection{DiffPure~\citep{Nie2022DiffusionMF}}
Given a forward diffusion process $\mathbf{x}(t)_{t\in[0, 1]}$, DiffPure uses $t^* = 0.1$ and $t^* = 0.075$ on CIFAR-10 against threat models $\ell_\infty (\epsilon = 8/255)$ and $\ell_2 (\epsilon = 0.5)$, respectively (with step size 0.001). Since it is impossible to calculate the gradients using back-propagation, in the original evaluation, DiffPure uses an adjoint method of its underlying numerical SDE (ODE) solver for calculating gradients. In our evaluation, we use gradients of a surrogate process calculated by direct back-propagation. To overcome memory constraints, we increase the step size of the surrogate denoising process in attack to 0.005.

\subsection{GDMP~\citep{Wang2022GuidedDM}}
The basic approach of GDMP is the same as DiffPure~\citep{Nie2022DiffusionMF}, however, it uses two additional techniques: guidance and multiple purification steps. GDMP proposes to use gradients of a distance between an initial input and a target being processed to preserve semantic information:
\begin{equation}
    \mathbf{x}_{t-1} \sim \mathcal{N}(\boldsymbol{\mu}_\theta - s\boldsymbol{\Sigma}_{\theta} \nabla_{\mathbf{x}_{t}} \mathcal{D}(\mathbf{x}_{t}, \mathbf{x}_{\text{adv}, t}), \boldsymbol{\Sigma}_{\theta}),
\end{equation}
where $\boldsymbol{\mu}_\theta$ and $\boldsymbol{\Sigma}_\theta$ are the mean and variance of $\mathbf{x}_{t-1}$ calculated by diffusion models $\boldsymbol{\epsilon}_\theta$, $s$ is a scale of guidance, $\mathbf{x}_{t}$ is a sample that is being purified, and $\mathbf{x}_{\text{adv}, t}$ is a noisy adversarial example. In addition, GDMP finds that iteratively applying the purification process, which we call the purification step, can improve the robustness. GDMP consists of four purification steps, each consisting of 36 forward steps and 36 denoising steps. In our evaluation, we use a surrogate process calculated by direct back-propagation, while GDMP is originally evaluated on BPDA. Since it is impossible to calculate the gradients of the full defense process, we use a surrogate process consisting of four purification steps (each consisting of 36 forward steps and six denoising steps) in the attack.

\subsection{SODEF and DISCO~\citep{kang2021stable, ho2022disco}}
SODEF~\citep{kang2021stable} uses an ODE block that satisfies Lyapunov stability. Lyapunov-stable equilibrium point has a property that its neighborhood gathers to that point by passing through the ODE block. In the original implementation, SODEF uses an adjoint method to calculate its gradients of the ODE block. In our experiment, we use back-propagation instead of the adjoint method.

DISCO~\citep{ho2022disco} employs a local implicit module to restore a clean example. For every pixel, the module estimates the original RGB values of input before sending it to the classifier. While DISCO is evaluated on BPDA in the original evaluation, we use the full gradients of the defense process. To evaluate both SODEF and DISCO, we use the standard version of AutoAttack.


\section{Additional Results on Evaluation for Diffusion-Based Purifications}
\label{app:evaluation}

\subsection{Additional Results on RQ1}
We show the difference between attack success rate when using the adjoint method and back-propagation. \autoref{table:RQ1_extra} shows the robust accuracy of DiffPure~\citep{Nie2022DiffusionMF} and its probability flow ODE~\citep{Song2020ScoreBasedGM} against PGD+EOT $\ell_\infty (\epsilon = 8/255)$ on CIFAR-10. For both DiffPure and its probability flow ODE, the back-propagation can generate adversarial examples more successfully than the adjoint method. For the probability flow ODE, the back-propagation with step size 0.01 has 34.51\% lower robust accuracy than the adjoint method.

\begin{table}[!ht]
    \centering
    \begin{tabular}{cccc}
    \toprule
        Defense & Method & Step Size in Attack & Robust Accuracy (\%) \\ \midrule
        \multirow{4}{*}{DiffPure~\citep{Nie2022DiffusionMF}} & Adjoint & 0.001 & 74.38±1.03 \\
        & Surrogate & 0.005 & 46.84±1.44 \\
        & Surrogate & 0.010 & 50.12±1.18 \\ 
        & Surrogate & 0.025 & 62.93±1.02 \\ \midrule
        \multirow{3}{*}{Probability flow ODE} & Adjoint & 0.010 & 70.47±0.99 \\
        & Full & 0.010 & 35.96±0.89 \\
        & Surrogate & 0.025 & 36.41±1.30 \\ \bottomrule
    \end{tabular}
    \caption{Robust accuracy of DiffPure and its probability flow ODE against PGD+EOT $\ell_\infty(\epsilon = 8/255)$ on CIFAR-10. We compare the attack success rate between the adjoint method and back-propagation. The gradient of the full or surrogate process is calculated by back-propagation. The maximum timestep $t^*$ is set to 0.1. The step size of DiffPure and probability flow ODE in defense is 0.001 and 0.01, respectively.}
    \label{table:RQ1_extra}
\end{table}   

\subsection{Additional Results on RQ2}

\begin{figure}[!ht]
    \centering
    \includegraphics[width=0.4\linewidth]{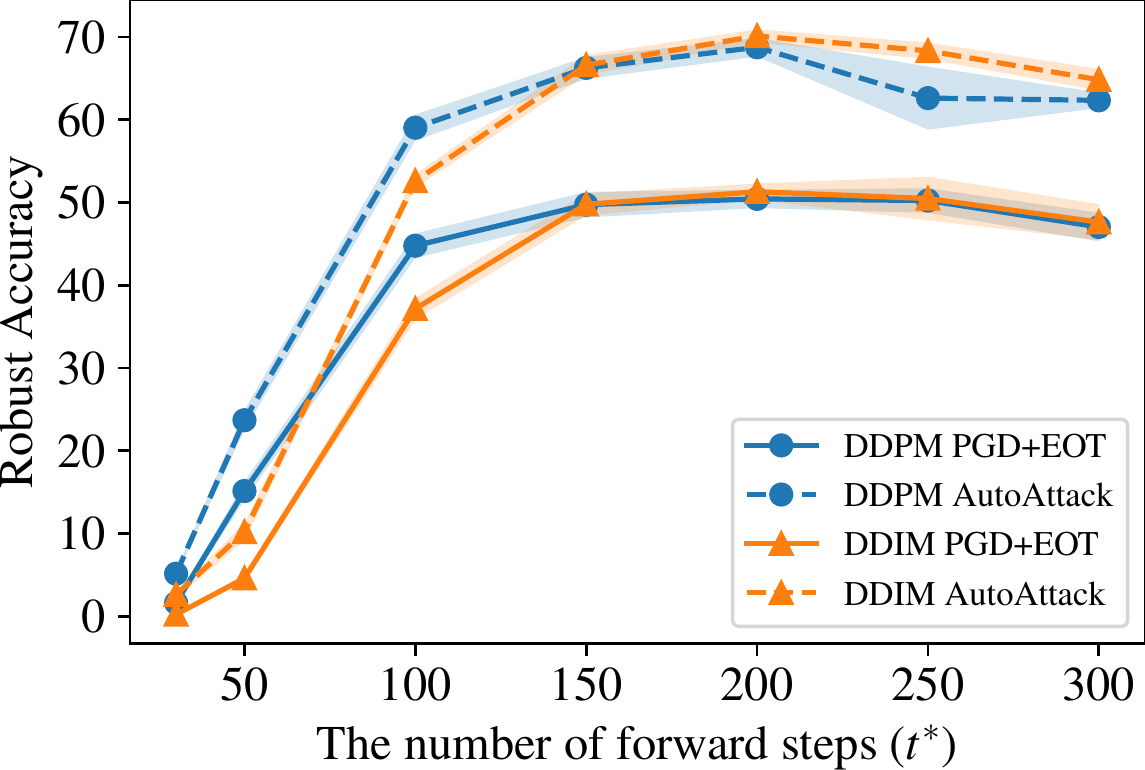}
    \caption{Standard and robust accuracy as we change the number of forward steps against PGD+EOT and AutoAttack $\ell_\infty (\epsilon = 8/255)$ on CIFAR-10. Five denoising steps for both attack and defense are used.}
    \label{fig:AA_vs_PGD}
\end{figure}

\autoref{fig:AA_vs_PGD} compares the attack performance of PGD and AutoAttack $\ell_\infty (\epsilon = 8/255)$ on CIFAR-10. This result is conducted on the same settings with \autoref{sec:hyperparameter}. As shown in the \autoref{fig:AA_vs_PGD}, PGD generally has a larger attack success rate than AutoAttack. As the number of forward steps increases, the gap between the two attacks also increases. Therefore, PGD+EOT is a more appropriate attack than AutoAttack for evaluating diffusion-based purification methods.

\section{Additional Results on Analysis of Hyperparameters}
\label{app:hyperparameter}

In this section, we provide further analyses of hyperparameters on CIFAR-10 and ImageNet. In all experiments, the implementation of diffusion-based purification follows noise scheduling, the forward process, and the denoising process of Appendix~\ref{app:diffusion_based_purification}.

\subsection{CIFAR-10}

\begin{figure}[!ht]
    \centering
    \includegraphics[width=\linewidth]
    {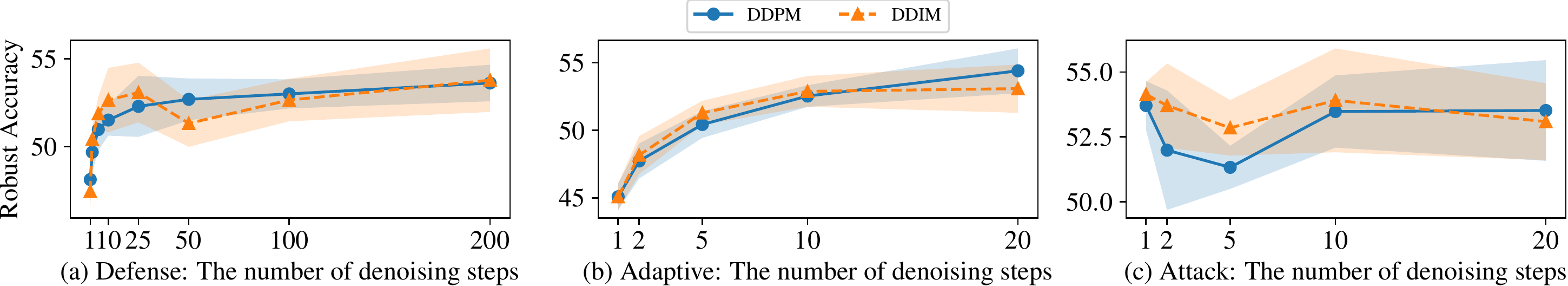}
    \caption{Robust accuracy as we change the number of denoising steps against PGD+EOT $\ell_\infty (\epsilon = 8/255)$ on CIFAR-10. We change the number of denoising steps in (a) defense, (b) both, and (c) attack for each experiment when the other hyperparameters are fixed. The number of forward steps is 200 (i.e., $t^* = 200$).}
    \label{fig:sampling_steps_t200}
\end{figure}

\begin{figure}[!ht]
   \begin{minipage}{0.48\textwidth}
     \centering
     \includegraphics[width=.8\linewidth]{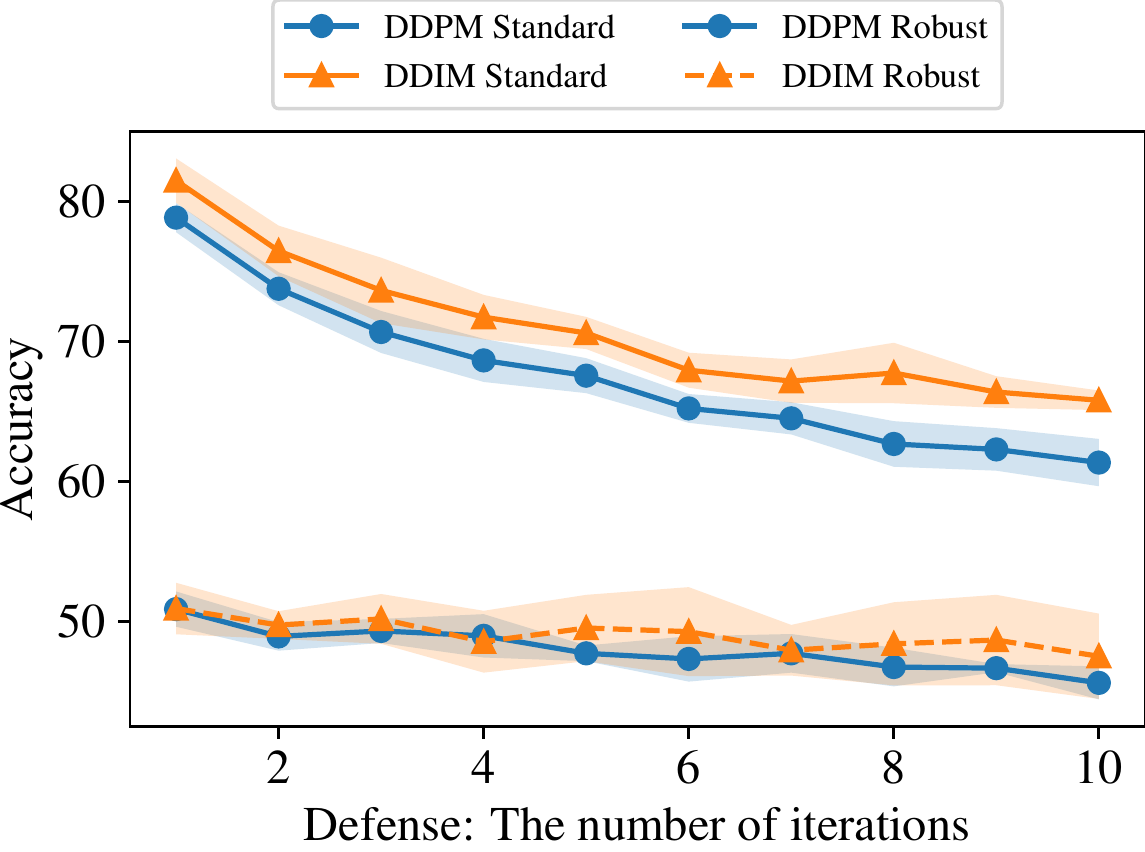}
    \caption{
    The number of purification steps in defense and its influence on the standard and robust accuracy against PGD+EOT $\ell_\infty (\epsilon = 8/255)$ on CIFAR-10. The number of forward steps is 200 (i.e., $t^* = 200$). The reported robust accuracy is the lowest performance among the various number of purification steps in the attacks.}
    \label{fig:defense_num_iterations_t200}
   \end{minipage}\hfill
   \begin{minipage}{0.48\textwidth}
     \centering
     \includegraphics[width=.8\linewidth]{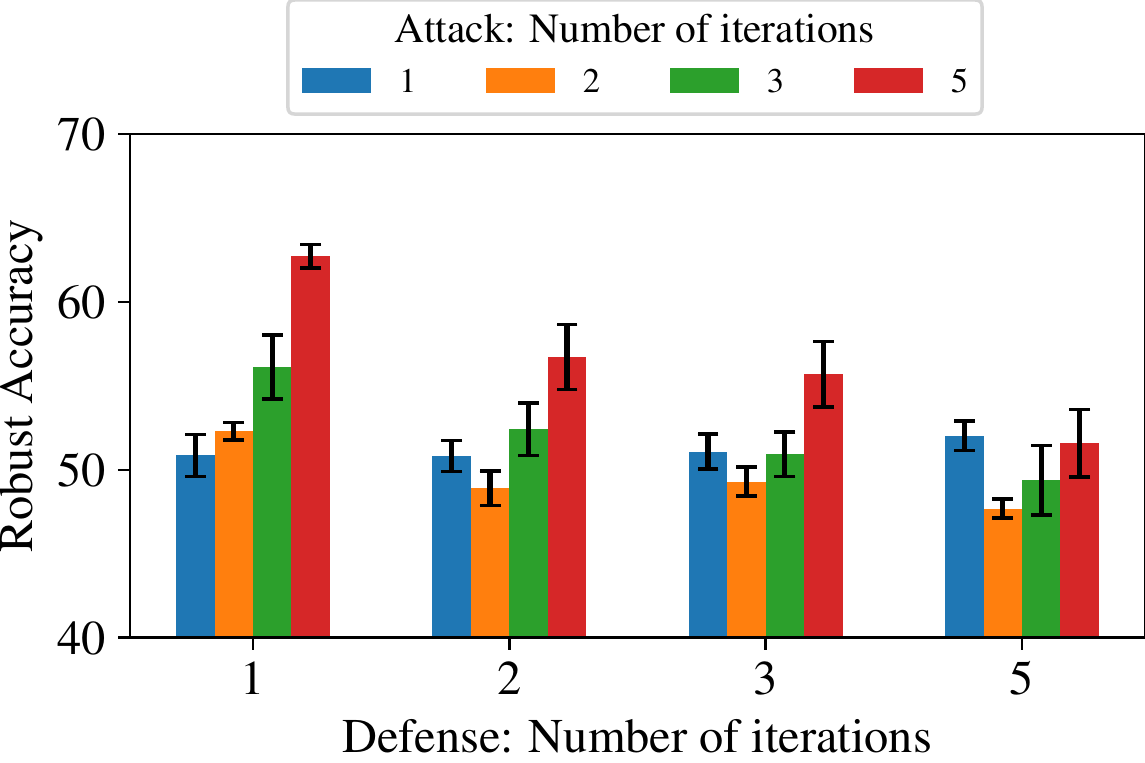}
    \caption{
    The number of purification steps during the attacks and its influence on the robust accuracy against PGD+EOT $\ell_\infty (\epsilon = 8/255)$ on CIFAR-10. The number of forward steps is 200 (i.e., $t^* = 200$)}.
    \label{fig:attack_num_iterations_t200}
   \end{minipage}
\end{figure}

\begin{figure}[!ht]
    \centering
    \includegraphics[width=\linewidth]{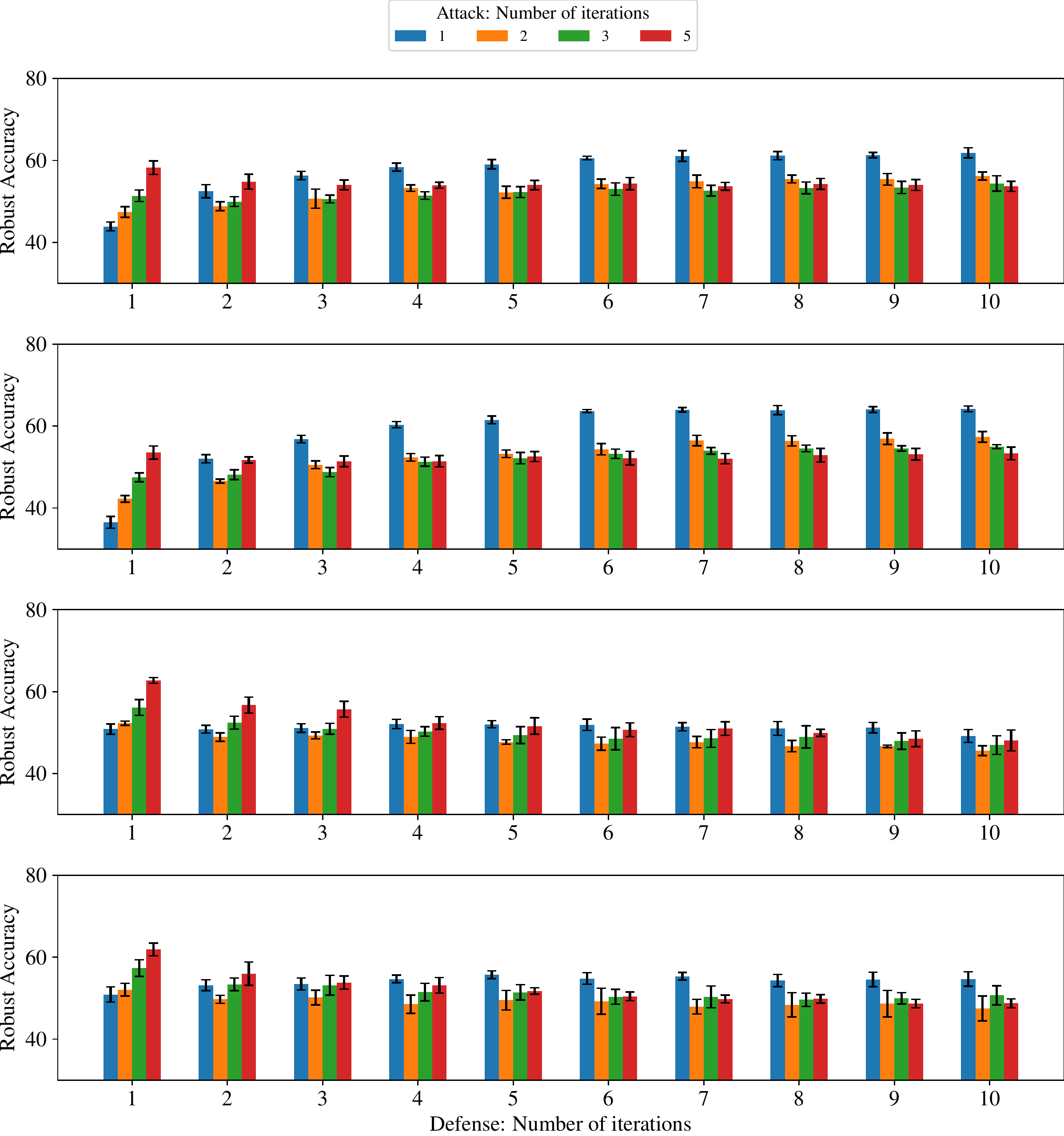}
    \caption{The number of purification steps during defenses and its influence on the robust accuracy against PGD+EOT $\ell_\infty (\epsilon = 8/255)$. We compare the attack success rate with respect to the number of purification steps during attacks. The number of forward steps of the top two rows is 100, and the number of forward steps of the bottom two rows is 200.}
    \label{fig:iterations_attack}
\end{figure}

We provide additional results regarding denoising steps and purification steps on CIFAR-10 through \autoref{fig:sampling_steps_t200}, \autoref{fig:defense_num_iterations_t200}, and \autoref{fig:attack_num_iterations_t200}. We conduct all experiments in a setting identical to \autoref{sec:denoising_steps}, except that the number of forward steps is 200 (i.e., $t^* = 200$). Furthermore, \autoref{fig:iterations_attack} shows the overall influence of the number of purification steps in attack on the attack success rate.


\subsection{ImageNet}

\begin{figure}[!ht]
    \centering
    \includegraphics[width=0.5\linewidth]{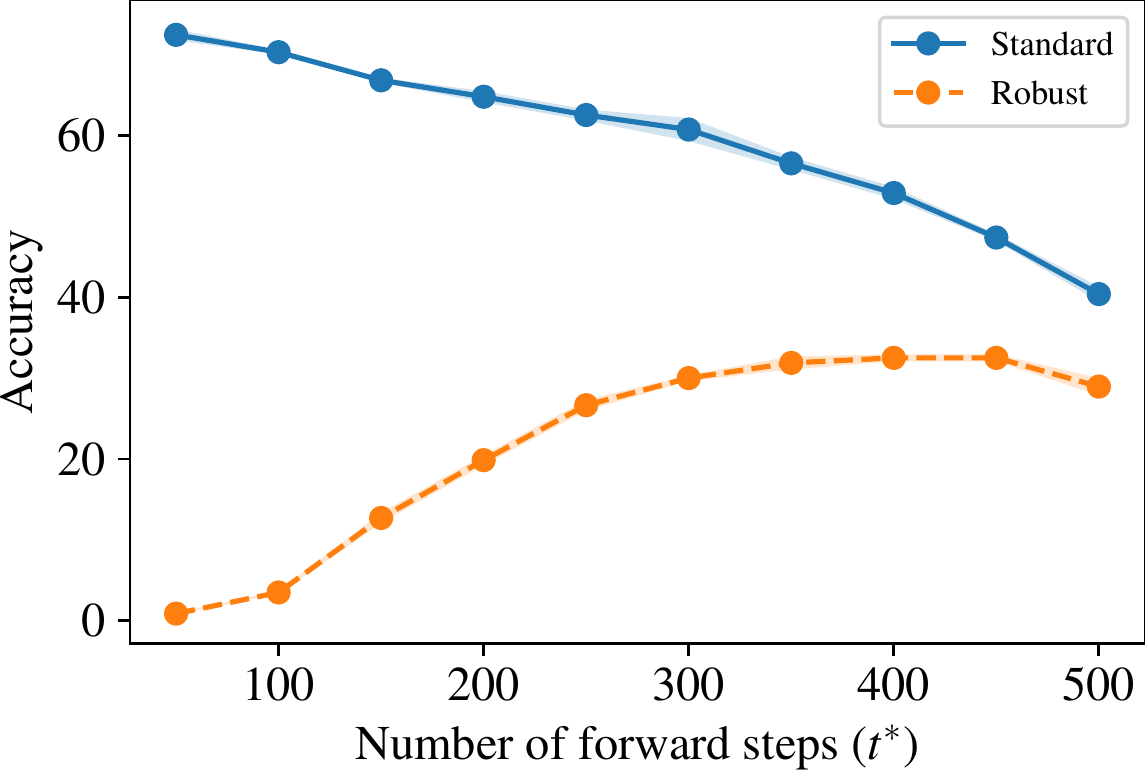}
    \caption{Standard and robust accuracy as we change the number of forward steps against PGD+EOT $\ell_\infty (\epsilon = 4/255)$ on ImageNet. Five denoising steps for both attack and defense are used. The change of total variance ranged from 0.03 to 0.9222.}
    \label{fig:imagenet_forward_steps}
\end{figure}

\begin{figure}[!ht]
    \centering
    \includegraphics[width=\linewidth]{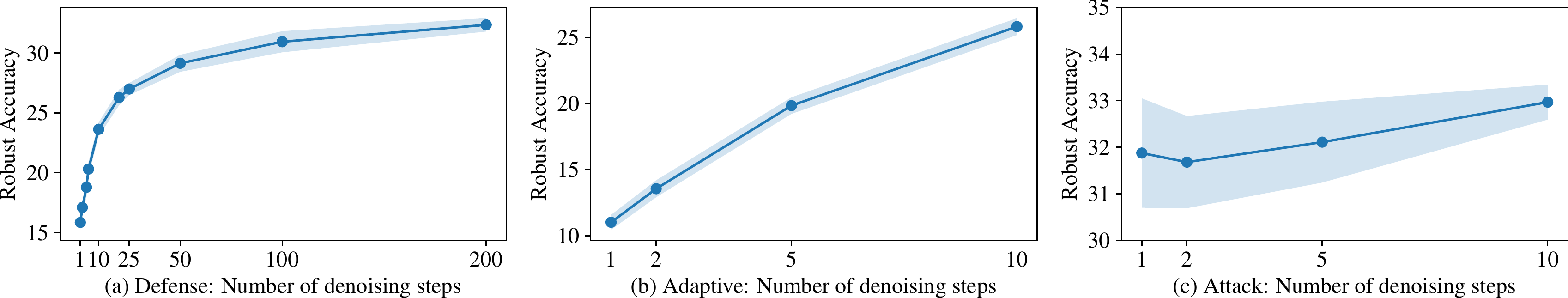}
    \caption{Robust accuracy as we change the number of denoising steps against PGD+EOT $\ell_\infty (\epsilon = 4/255)$ on ImageNet. We change the number of denoising steps in (a) defense, (b) both, and (c) attack for each experiment when the other hyperparameters are fixed. The number of forward steps is 200 (i.e., $t^* = 200$).}
    \label{fig:imagenet_sampling_steps_t200}
\end{figure}

We provide additional results regarding forward, denoising, and purification steps on ImageNet through \autoref{fig:imagenet_forward_steps}, \autoref{fig:imagenet_sampling_steps_t200}, and \autoref{table:imagenet_iterations}, respectively. In \autoref{fig:imagenet_sampling_steps_t200} and \autoref{table:imagenet_iterations}, the number of forward steps is set to 200. For ImageNet, we use 20 PGD iterations and 20 EOT samples. And due to memory constraints, the upper bound on the number of function calls is set to ten. Although we employ the experiments only on DDPM, the results are similar to those from CIFAR-10.

\begin{table}[!ht]
    \centering
    \begin{tabular}{cccc}
    \toprule
        \multirow{2}{*}{\# purification steps in defense} & \multirow{2}{*}{\# purification steps in attack} & \multicolumn{2}{c}{Accuracy (\%)} \\
        & & Standard & Robust \\ \midrule
        \multirow{2}{*}{1} & 1 & \multirow{2}{*}{64.80±0.70} & 19.84±0.63 \\
        & 2 & & 26.84±0.45 \\
        \multirow{2}{*}{2} & 1 & \multirow{2}{*}{59.06±0.92} & 27.85±0.71 \\
         & 2 & & 25.00±1.13 \\ \bottomrule
    \end{tabular}
    \caption{Standard accuracy and Robust accuracy against PGD+EOT $\ell_\infty (\epsilon = 4/255)$ on ImageNet. We compare the accuracy between the different number of purification steps in attack and defense.}
    \label{table:imagenet_iterations}
\end{table}         


\section{Surrogate Process for Gradual Noise-Scheduling}
\label{app:gradual_sampling}
The defense process of the gradual noise-scheduling have three levels in the number of forward steps. To validate the robustness of our defense, we evaluate the gradual noise-scheduling with several surrogate processes and report the lowest robust accuracy among them. We denote one purification step as (\# of forward steps, \# of denoising steps). For example, suppose a process uses two purification steps, consisting of 100 forward steps with 20 denoising steps and 120 forward steps with ten denoising steps, respectively. In that case, we denote the defense process as (100, 20), (120, 10). As shown in \autoref{table:attack_processes}, the third surrogate process has the lowest robust accuracy against the gradual noise-scheduling on CIFAR-10. For all experiments of our defense on all datasets, we select the third process as the surrogate process in attacks of the adaptive white-box setting.

\begin{table}[!ht]
    \centering
    \resizebox{.5\linewidth}{!}{
        \begin{tabular}{lc}
        \toprule
            Attack Process & Robust Accuracy (\%) \\ \midrule
            (30, 5), (50, 5), (125, 5), (125, 5) & 56.80±1.12 \\
            (30, 1), (50, 1), (125, 10) & 56.13±1.04 \\
            (30, 1), (50, 1), (125, 5) & 55.82±0.59 \\
            (125, 5), (125, 5) & 62.73±0.74 \\
            (125, 10) & 60.16±1.02 \\
            (125, 5) & 61.60±0.63 \\ \bottomrule
        \end{tabular}
    }
    \caption{Robust accuracy against PGD+EOT $\ell_\infty (\epsilon = 8/255)$ on CIFAR-10. The attack processes are used for generating adversarial examples against our defense strategy.}
    \label{table:attack_processes}
\end{table}   


\section{Memory and Time Requirements for Diffusion-Based Purification}

\begin{figure}[!ht]
    \centering
    \includegraphics[width=.8\linewidth]
    {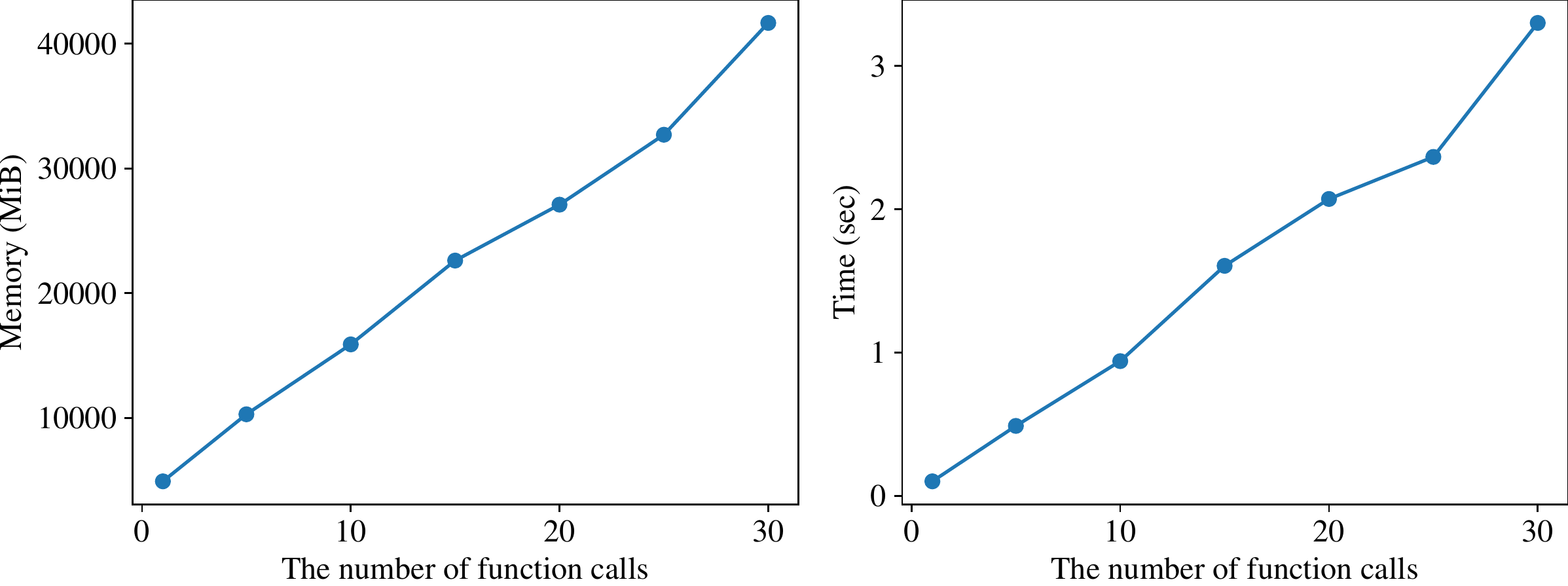}
    \caption{Memory and time usage of diffusion-based purification on CIFAR-10. We use eight for batch size.}
    \label{fig:memory_and_time}
\end{figure}

Evaluating diffusion-based purification requires lots of memory since we use direct back-propagation. \autoref{fig:memory_and_time} briefly shows the memory and time requirements of one PGD iteration for implementing diffusion-based purification methods. We use one A100 GPU. For example, we need almost ten days to evaluate one experiment with 30 function calls for calculating back-propagation with one A100 GPU (against a PGD attack using 200 iterations and 20 EOT samples).

\end{document}